\documentclass[a4paper,fleqn]{cas-dc}

\usepackage[numbers]{natbib}
\usepackage{amsmath,amssymb}
\usepackage{algorithm}
\usepackage{algpseudocode}
\usepackage{booktabs}
\usepackage{multirow}
\usepackage{makecell}
\usepackage[section]{placeins}
\usepackage{array}
\usepackage{graphicx}
\usepackage{pifont}
\usepackage{stfloats}
\usepackage{float}

\begin{document}
\let\WriteBookmarks\relax
\def\floatpagepagefraction{1}
\def\textpagefraction{.001}

\shorttitle{Granular Ball Guided Stable Latent Domain Discovery for DG Crowd Counting}

\shortauthors{F. Chen et al.}

% Main title
\title[mode=title]{Granular Ball Guided Stable Latent Domain Discovery for Domain-General Crowd Counting}

% Title footnote
\tnotemark[1]
\tnotetext[1]{This work was supported in part by the National Natural Science Foundation of China under Grant Nos. 62221005, Chongqing Natural Science Foundation Innovation and Development Joint Fund No. CSTB2025NSCQ-LZX0141.}

% Authors
\author[1]{Fan Chen}[orcid=0009-0007-2163-5557,bioid=1]
\author[1]{Shuyin Xia}[orcid=0000-0001-5993-9563,bioid=2]
\cormark[1]
\ead{xiasy@cqupt.edu.cn}

\author[2]{Yi Wang}[bioid=3]

% Affiliations
\affiliation[1]{organization={Chongqing University of Posts and Telecommunications},
                city={Chongqing},
                country={China}}

\affiliation[2]{organization={Chongqing Ant Consumer Finance Co., Ltd.},
                city={Chongqing},
                country={China}}

\cortext[1]{Corresponding author.}

\begin{abstract}
Single-source domain generalization for crowd counting is highly challenging because a single labeled source domain may contain heterogeneous latent domains, while unseen target domains often exhibit severe distribution shifts. A central issue is stable latent domain discovery: directly performing flat clustering on evolving sample-level latent features is easily disturbed by feature noise, outliers, and representation drift, leading to unreliable pseudo-domain assignments and weakened domain-structured learning. To address this problem, we propose a granular ball guided stable latent domain discovery framework for domain-general crowd counting. The proposed method first groups samples into compact local granular balls and then clusters granular ball centers as representatives to infer pseudo-domains, thereby converting direct sample-level clustering into a hierarchical representative-based clustering process. This design produces more stable and semantically consistent pseudo-domain assignments. On top of the discovered latent domains, we develop a two-branch learning framework that improves transferable semantic representations via semantic codebook re-encoding and captures domain-specific appearance variations through a style branch, thereby alleviating semantic--style entanglement under domain shifts. Extensive experiments on ShanghaiTech A/B, UCF\_QNRF, and NWPU-Crowd under a strict no-adaptation protocol verify the effectiveness of the proposed method and show strong generalization ability, especially in transfer settings with large domain gaps.
\end{abstract}

% \begin{highlights}
% \item We propose a granular ball guided framework for stable latent domain discovery in single-source domain-general crowd counting.
% \item Granular ball centers are clustered as representatives to obtain more stable pseudo-domains than direct sample-level flat clustering.
% \item A two-branch design with semantic codebook re-encoding and structured branch regularization improves cross-domain counting generalization.
% \end{highlights}

\begin{keywords}
crowd counting \sep domain generalization \sep single-source domain generalization \sep stable latent domain discovery \sep granular ball \sep disentangled representation learning
\end{keywords}

\maketitle

\section{Introduction}
\label{sec:intro}

Crowd counting aims to estimate the number of people in an image, typically by predicting a density map whose integral yields the final count~\cite{zhang2016single,li2018csrnet,wang2020distribution,cao2018scale,liu2019context,ma2019bayesian,yi2023effective,10699432}.
It plays a critical role in public safety monitoring, large-scale event management, and intelligent transportation.
Recent advances have been largely driven by deep neural networks. However, most counting models rely on the assumption that the deployment data distribution matches the training distribution, and their performance can drop drastically when deployed to data distributions different from those seen during training~\cite{zhang2016single,li2018csrnet,wang2022generalizing}. In real-world applications, crowd images collected across different cities, cameras, and time periods often exhibit substantial domain shifts, including changes in density levels, viewpoints, scene layouts, illumination, and background clutter~\cite{idrees2018composition,wang2020nwpu}. Such shifts are particularly challenging for density regression, as accurate counting relies on both semantic cues and fine-grained texture and scale patterns~\cite{li2018csrnet,cao2018scale,wang2020distribution}. Meanwhile, annotating dense point labels or density maps for every new environment is expensive and sometimes infeasible~\cite{gao2021domain,reddy2021adacrowd}. A common way to address domain shift is domain adaptation, which leverages unlabeled or sparsely labeled target data to reduce distribution gaps~\cite{ganin2016domain,sun2016deep}. However, target data may be unavailable due to privacy constraints, limited access, or fast-changing deployment environments. This motivates domain generalization (DG): learning a counting model from a labeled source domain only, while generalizing to arbitrary unseen target domains without any further adaptation~\cite{wang2022generalizing,gulrajani2020search,zhou2021domain}.

\begin{figure}[]
    \centering
    \includegraphics[width=\linewidth]{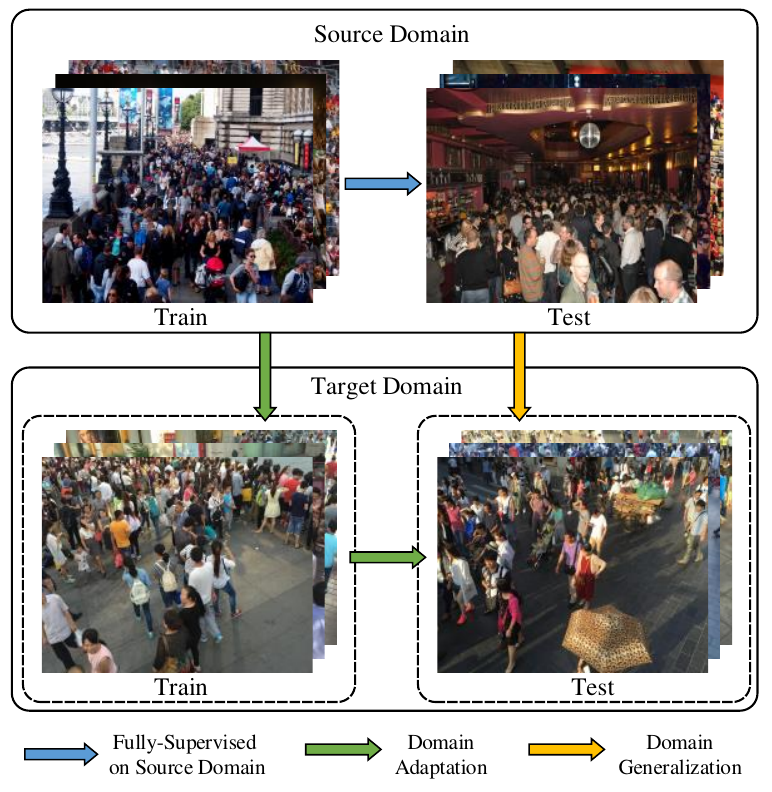}
    \caption{Illustration of three settings under domain shift: fully supervised learning, domain adaptation, and domain generalization.}
    \label{fig:motivation}
\end{figure}

As illustrated in Fig.~\ref{fig:motivation}, existing counting pipelines under distribution shift can be roughly categorized into three settings.
Fully-supervised methods assume that training and test images come from the same source domain, and thus performance deteriorates when the test distribution shifts.
Domain adaptation alleviates this issue by exploiting unlabeled (or sparsely labeled) target-domain training data to align distributions before testing, but it explicitly requires access to target data during training.
In contrast, domain generalization targets a more practical deployment scenario: the model is trained only on the labeled source domain and is expected to generalize directly to arbitrary unseen target domains at test time, without any target-domain data or further adaptation.

A key challenge of single-source DG is that a ``source domain'' is rarely homogeneous. Instead, it usually contains multiple latent domains induced by diverse scenes, imaging conditions, and crowd patterns. Therefore, domain-general crowd counting is not only a problem of learning domain-invariant representations, but also a problem of reliably discovering latent domain structure hidden within the source set. Recent DG approaches attempt to partition the source data into pseudo-domains and exploit them to enhance robustness under domain shifts~\cite{du2023domain,ding2024domain}. Some recent works further leverage predictive uncertainty to guide style diversification for improving generalization~\cite{ding2024domain,gal2016dropout,kendall2017uncertainties,lakshminarayanan2017simple}. However, a fundamental issue remains under-explored: stable latent domain discovery in an evolving feature space. Existing pseudo-domain discovery methods commonly rely on sample-level flat clustering over learned features, which is easily affected by noisy samples, outliers, and representation drift during training, resulting in unstable pseudo-domain assignments~\cite{du2023domain}. Moreover, even when pseudo-domains are available, it is still non-trivial to learn counting representations that preserve counting-relevant semantics while preventing domain-specific appearance factors from leaking into the shared representation~\cite{du2023domain,ding2024domain}.

Granular Ball Computing (GBC) provides a promising perspective for addressing this challenge. As a lightweight paradigm for structural multi-granularity modeling, GBC organizes samples into compact granular balls that summarize local structure with ball-level representatives, and progressively refines the partition from coarse to fine~\cite{wang2017dgcc,Xia2023GranularballCA,xie2024gbg++,xie2024w}. This property is particularly appealing for latent domain discovery in single-source DG crowd counting. Compared with directly clustering individual samples, granular ball partitioning first aggregates locally consistent samples into compact structured groups and then uses granular ball centers as stable representatives for higher-level pseudo-domain grouping, thereby reducing sensitivity to noisy features and transient representation drift. In this way, latent domain discovery can be transformed from unstable sample-level flat clustering into a more robust hierarchical representative-based clustering process. Meanwhile, recent GBC studies have demonstrated its effectiveness in visual understanding and representation modeling, suggesting that granular ball structures can serve as efficient and interpretable units for adaptive multi-granularity computation~\cite{cheng2024gb,quadir2024granular,10996538}.

Motivated by the above observations, we propose Granular Ball Guided Stable Latent Domain Discovery for domain-general crowd counting. Our key idea is to model the latent domain structure of a single source dataset from two complementary perspectives: stable structure discovery and structure-aware representation learning. First, we introduce a granular ball guided latent domain discovery strategy in a domain-sensitive feature space. Instead of directly clustering individual samples, we recursively partition the feature set into compact granular balls and then cluster granular ball centers as ball-level representatives to obtain pseudo-domains. This coarse-to-fine and set-to-representative procedure transforms unstable sample-level flat clustering into a more robust hierarchical representative-based clustering process, thereby improving the stability of pseudo-domain assignments and enabling online refinement during training. Second, based on the discovered latent domains, we build a two-branch learning scheme on top of a shared multi-scale backbone. The semantic branch employs a learnable semantic codebook to re-encode semantic features into a compact and stable basis for density estimation, while the style branch captures domain-related appearance variations and provides regularization signals. By enforcing intra-domain style compactness and semantic--style orthogonality, the proposed framework reduces domain leakage and facilitates disentangled counting representation learning, leading to improved robustness under domain shifts~\cite{du2023domain,ding2024domain}.

We evaluate our granular ball guided stable latent domain discovery framework for domain-general crowd counting, referred to as GBDGC, on standard DG crowd counting benchmarks and demonstrate consistent improvements over strong baselines across multiple cross-domain transfer settings. In particular, GBDGC yields clear gains on challenging transfers such as SHB$\rightarrow$QNRF, highlighting the effectiveness of granular ball guided stable latent domain discovery and semantic codebook re-encoding under large domain gaps.

Our main contributions are summarized as follows:
\begin{itemize}
    \item We identify stable latent domain discovery as a key bottleneck in single-source domain-general crowd counting, and show that existing sample-level flat clustering strategies are vulnerable to noisy features, outliers, and representation drift in evolving latent spaces.
    \item We propose a granular ball guided stable latent domain discovery method that first partitions samples into compact granular balls and then clusters granular ball centers as representatives, thereby transforming direct sample-level clustering into a hierarchical representative-based clustering process and yielding more stable pseudo-domain assignments.
    \item We develop a structure-aware counting framework built upon the discovered latent domains, with semantic codebook re-encoding and structured branch regularization, which enhances domain-invariant semantic learning while suppressing domain-specific appearance leakage, and thus improves generalization to unseen domains.
\end{itemize}

The remainder of this paper is organized as follows.
Section~\ref{sec:rw} reviews the related work.
Section~\ref{sec:method} details the proposed method.
Section~\ref{sec:exp} presents the experimental results and ablation studies.
Finally, Section~\ref{sec:conclusion} concludes the paper.

\section{Related Work}
\label{sec:rw}

\subsection{Crowd Counting}
Crowd counting aims to estimate the number of people in an image, typically by predicting a density map whose integral yields the final count. Early deep counting methods mainly addressed severe scale variation through multi-column or multi-branch architectures, among which MCNN~\cite{zhang2016single} is a representative example. Subsequent studies improved counting performance by enhancing feature representation and context modeling, such as CSRNet~\cite{li2018csrnet}, SANet~\cite{cao2018scale}, and other context-aware or scale-aggregation frameworks~\cite{liu2019context}. These methods significantly advanced density regression under congested scenes and complex backgrounds. More recently, transformer-based models and more discriminative learning paradigms have been introduced to strengthen global dependency modeling and scene understanding in crowd counting~\cite{mo2024countformer,chen2024learning}. In parallel, a series of works have revisited the supervision and optimization side of counting, including composition loss~\cite{idrees2018composition}, Bayesian loss for point-supervised learning~\cite{ma2019bayesian}, and distribution-based objectives that reduce the bias introduced by Gaussian-smoothed labels~\cite{wang2020distribution}. Large-scale benchmarks such as NWPU-Crowd~\cite{wang2020nwpu} further promote evaluation under diverse scenes and density levels. Despite these advances, most existing counting models are still developed under the assumption that training and test data follow similar distributions. When deployed across domains with substantial shifts in viewpoint, density range, illumination, and background clutter, their performance often degrades markedly. This limitation motivates the study of domain generalization for crowd counting.

\subsection{Domain Generalization}
Cross-domain crowd counting is challenging because the target domain may differ significantly from the source domain in both imaging conditions and crowd characteristics. Existing efforts mainly follow two related paradigms: domain adaptation and domain generalization. Domain adaptation assumes access to target-domain data during training and reduces distribution discrepancy through adversarial alignment, feature transfer, or consistency regularization~\cite{ganin2016domain,sun2016deep,gao2021domain,reddy2021adacrowd,zhu2023daot,xie2023striking}. While effective, such methods are less practical when target data are unavailable, privacy-sensitive, or continuously changing. Domain generalization (DG), in contrast, requires the model to generalize to unseen domains using only source-domain data. General DG methods often improve robustness through style randomization, feature normalization, regularization, or meta-learning strategies~\cite{wang2022generalizing,gulrajani2020search,zhou2021domain}. In crowd counting, recent single-source DG methods attempt to uncover latent sub-domains within the source dataset and exploit pseudo-domain supervision to improve generalization~\cite{du2023domain,peng2024single}. Related studies also introduce style diversification, uncertainty estimation, or augmentation-based mechanisms to improve robustness under distribution shifts~\cite{ding2024domain,gal2016dropout,kendall2017uncertainties,lakshminarayanan2017simple}. However, existing single-source DG counting methods still largely rely on sample-level flat clustering over evolving feature spaces. Such strategies are sensitive to noisy samples, outliers, and representation drift, which may lead to unstable pseudo-domain assignments. Moreover, current approaches remain limited in simultaneously preserving counting-relevant semantics and suppressing domain-specific appearance variations. These limitations motivate us to seek a more stable latent structure discovery mechanism for single-source DG crowd counting.

\subsection{Granular Ball Computing}
Granular Ball Computing (GBC) is a multi-granularity data representation paradigm that models samples by a set of adaptive granular balls with different sizes, where each ball summarizes a local region through representative statistics such as center and radius~\cite{Xia2023GranularballCA,xia2022efficient}. By progressively refining coarse partitions into finer ones, GBC captures hierarchical data structure while maintaining robustness to local perturbations and noise. Owing to its efficiency, interpretability, and tolerance to heterogeneous data distributions, GBC has been explored in classification, clustering, and other learning tasks~\cite{xia2020ball,xia2022efficient,xie2024mgnr}. These properties make GBC particularly attractive for latent structure discovery in domain generalization. Instead of directly clustering individual samples in a noisy and evolving feature space, GBC first aggregates nearby samples into compact local sets and then performs higher-level grouping over ball-level representatives. Such a hierarchical representative-based process is potentially more stable than flat sample-level clustering, since local noise and outliers can be absorbed before global grouping is performed. Therefore, GBC provides a natural structural perspective for pseudo-domain discovery in single-source DG crowd counting. To the best of our knowledge, granular ball based structural modeling has not been explicitly explored for stable latent domain discovery in crowd counting. This motivates us to introduce granular ball guided pseudo-domain partitioning and integrate it with structure-aware counting representation learning.

\section{Method}
\label{sec:method}

\begin{figure*}[!ht]
    \centering
    \includegraphics[width=\textwidth]{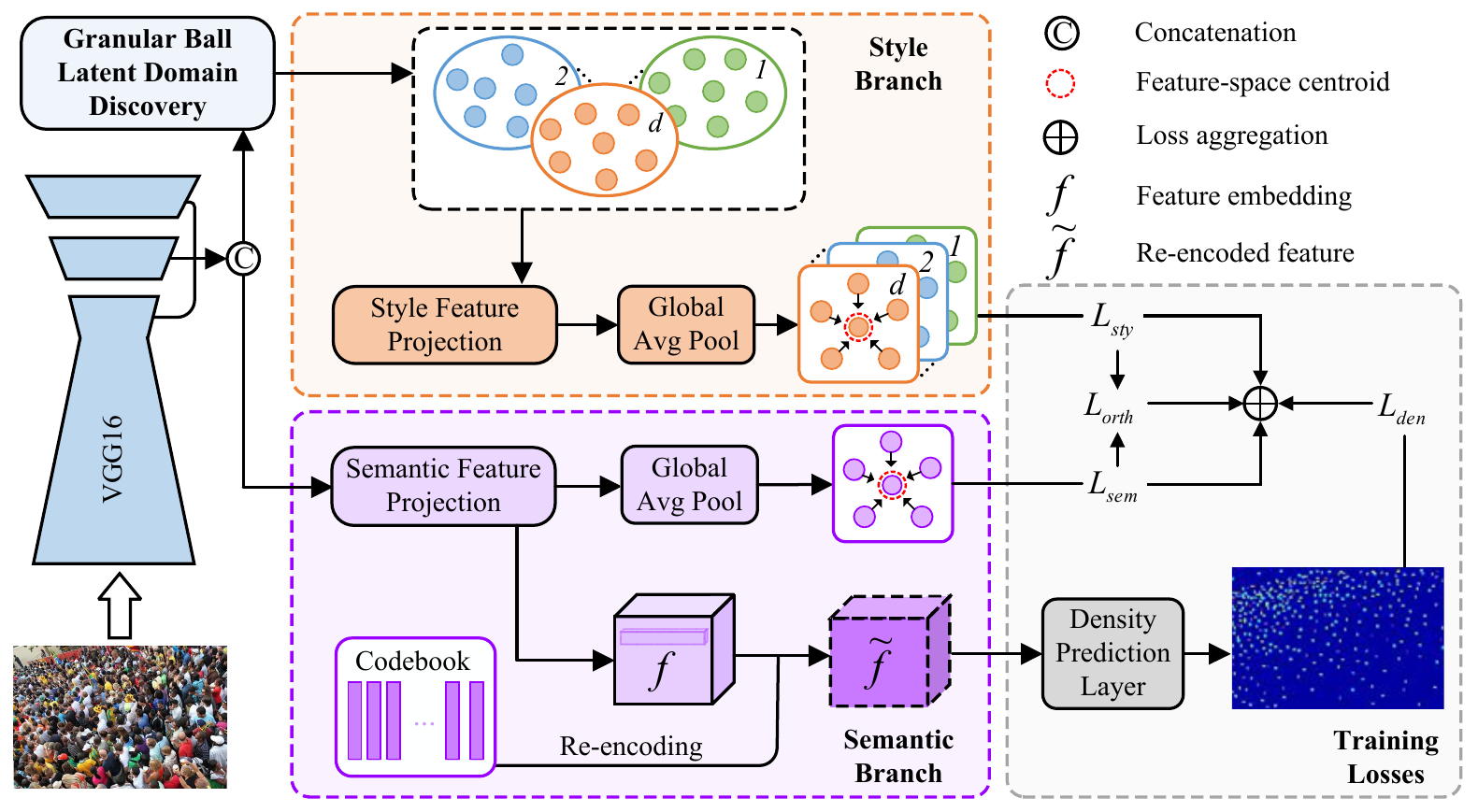}
    \caption{Overall framework of the proposed method, consisting of semantic codebook re-encoding, style-branch regularization, and granular ball guided stable latent domain discovery for robust density regression.}
    \label{fig:framework}
\end{figure*}

Granular Ball Computing (GBC) represents data by a set of compact local structural units, where each granular ball summarizes a subset of samples with a representative center and associated structural attributes. By recursively splitting balls from coarse to fine, GBC provides a hierarchical multi-granularity partition of the feature space. In this work, we adopt this idea for latent domain discovery, where granular balls serve as stable local structural units for grouping evolving sample representations. Given a dataset $X=\{x_1,x_2,\ldots,x_n\}$, we construct a granular ball set $GB=\{GB_1,GB_2,\ldots,GB_m\}$, where each ball is defined as $GB_i=(X_i,c_i,\vec{\theta}_i)$ with covered subset $X_i\subset X$, center $c_i$, and structural attributes $\vec{\theta}_i$. The general GBC model can be written as:
\begin{equation}\label{equ:GBC}
	\begin{aligned}
		&\quad \quad \quad \quad f(X,\vec{\alpha}) \longrightarrow  g(GB,\vec{\beta})\\
		&\min_{c_i,\vec{\theta}_{i},\vec{\beta},m} \ \mathcal{J} \big(-Cov(GB),-Comp(GB)\big) + m \\
		&\text{s.t.}\quad quality(GB_i) \geq \phi(X),\quad i = 1,2,\ldots,m \\
		&\qquad\ \ \ \  C(GB,X)\geq  0,
	\end{aligned}
\end{equation}
which reflects a transition from point-wise learning to multi-granularity ball-based computation.

As illustrated in Fig.~\ref{fig:framework}, the proposed method learns robust counting representations by explicitly modeling the latent domain structure hidden in a single source dataset. Given an input image, a VGG16-based encoder--decoder backbone extracts multi-level features, and three feature maps from different scales are upsampled and concatenated into a unified representation $F$. Based on $F$, we adopt a two-branch design to separate semantic and style factors. The semantic branch focuses on domain-invariant crowd semantics and re-encodes compact semantic features with a learnable codebook for density regression. In contrast, the style branch captures domain-related appearance variations and provides regularization signals to facilitate disentanglement and reduce semantic--style leakage.

To introduce latent domain structure without domain labels, we instantiate Eq.~(\ref{equ:GBC}) on domain-sensitive feature statistics $\mathbf{z}_i$ computed from the means and standard deviations of multi-level features. Instead of directly clustering individual samples, we first partition the evolving latent features into compact granular balls and then cluster granular ball centers as representatives to obtain $K$ pseudo-domains. This transforms direct sample-level flat clustering into a hierarchical representative-based clustering process, improving the stability of pseudo-domain assignments during training. Based on the resulting pseudo-domains, we compute domain-wise centers and enforce domain-structured constraints for robust representation learning. During inference, the style branch is discarded, and density maps are predicted using only the semantic branch with codebook re-encoding. In the following, we detail the semantic codebook re-encoding, style-branch regularization, and granular ball guided stable latent domain discovery modules.

\subsection{Semantic Codebook Re-encoding}
\label{sec:sem_codebook}

The semantic branch aims to learn stable counting semantics that are shared across the discovered latent domains. Given the fused feature $F$, we first project it with a $1\times1$ convolution to obtain a semantic feature map $S\in\mathbb{R}^{H'\times W'\times d}$. To enhance semantic compactness and improve robustness to domain-specific appearance variations, we introduce a learnable semantic codebook $E=[e_1,\ldots,e_{M_c}]\in\mathbb{R}^{d\times M_c}$ and re-encode each spatial location by soft assignment. Let $s_j\in\mathbb{R}^{d}$ denote the feature vector at spatial location $j$. We compute the assignment weights as
\begin{equation}
a_j=\mathrm{Softmax}\!\left(\frac{E^\top s_j}{\sqrt{d}}\right)\in\mathbb{R}^{M_c},
\end{equation}
and obtain the re-encoded semantic feature
\begin{equation}
\tilde{s}_j = E\,a_j \in \mathbb{R}^{d}.
\end{equation}
Stacking all $\{\tilde{s}_j\}$ yields the refined semantic feature map $\tilde{S}$, which is then fed into the density regressor to predict the density map $\hat{y}$. During inference, we keep only the semantic branch with codebook re-encoding for density estimation.

To further encourage consistency of semantic representations across pseudo-domains, we align image-level semantic descriptors derived from the re-encoded features. For each image, we compute the global semantic descriptor $p_i=\mathrm{GAP}(\tilde{S}_i)\in\mathbb{R}^{d}$, and denote the batch indices belonging to pseudo-domain $k$ as $\mathcal{I}_k$. Let
\begin{equation}
\nu_k=\frac{1}{|\mathcal{I}_k|}\sum_{i\in\mathcal{I}_k}p_i,\qquad
\nu=\frac{1}{B}\sum_{i=1}^{B}p_i,
\end{equation}
where $\nu_k$ is the semantic center of pseudo-domain $k$ within the mini-batch, and $\nu$ is the global batch semantic center. We minimize
\begin{equation}
\mathcal{L}_{\mathrm{sem}}
=\frac{1}{|\mathcal{K}_B|}\sum_{k\in\mathcal{K}_B}\|\nu_k-\nu\|_2^2,
\end{equation}
where $\mathcal{K}_B=\{k\mid|\mathcal{I}_k|>0\}$. This objective encourages different pseudo-domains to share a more consistent semantic basis, thereby enhancing domain-invariant counting representation learning.

\subsection{Structured Branch Regularization}
\label{sec:style}

The style branch is designed to model domain-specific appearance variations based on the discovered pseudo-domains. Given the shared multi-scale feature $F$, we apply a lightweight $1\times1$ convolutional projection in parallel to the semantic branch, and obtain a style feature map
\begin{equation}
T \in \mathbb{R}^{H'\times W'\times d}.
\end{equation}
Unlike the semantic branch, $T$ is not used for density estimation. Instead, it provides domain-structured regularization signals that help separate domain-related appearance factors from counting-relevant semantics.

Given the pseudo-domain labels $\{\hat d_i\}_{i=1}^{B}$ obtained by the stable latent domain discovery module, we define the index set of samples assigned to pseudo-domain $k$ as
$\mathcal{I}_k=\{ i \mid \hat d_i=k \}$, and denote the pseudo-domains appearing in the current batch by
$\mathcal{K}_B=\{k \mid |\mathcal{I}_k|>0\}$.
We summarize the style feature of each image by global average pooling and compute the corresponding pseudo-domain style center as
\begin{equation}
t_i=\mathrm{GAP}(T_i)\in\mathbb{R}^{d},\qquad
\bar t_k=\frac{1}{|\mathcal{I}_k|}\sum_{i\in\mathcal{I}_k} t_i.
\end{equation}
To encourage style consistency within each pseudo-domain, we minimize the intra-domain variance:
\begin{equation}
\mathcal{L}_{\mathrm{sty}}
=
\frac{1}{|\mathcal{K}_B|}\sum_{k\in\mathcal{K}_B}
\frac{1}{|\mathcal{I}_k|}\sum_{i\in\mathcal{I}_k}
\left\|t_i-\bar t_k\right\|_2^2.
\end{equation}
This objective encourages samples belonging to the same pseudo-domain to share more compact appearance statistics, thereby providing cleaner domain-wise style supervision for disentangled representation learning.

To further reduce information leakage between semantic and style representations, we impose an orthogonality constraint at the spatial level. Let $S$ and $T$ denote the pre-encoded semantic and style feature maps before codebook re-encoding, respectively, and flatten them as
$S^{\mathrm{flat}} \in \mathbb{R}^{d\times H'W'}$ and $T^{\mathrm{flat}} \in \mathbb{R}^{d\times H'W'}$.
We penalize the squared cosine similarity between the corresponding spatial vectors:
\begin{equation}
\mathcal{L}_{\mathrm{orth}}
= \frac{1}{H'W'}\sum_{j=1}^{H'W'}
\left(
\frac{ \tau_j^\top s_j }{ \|\tau_j\|_2\,\|s_j\|_2 + \epsilon }
\right)^2,
\end{equation}
where $s_j$ and $\tau_j$ are the $j$-th column vectors of $S^{\mathrm{flat}}$ and $T^{\mathrm{flat}}$, respectively, and $\epsilon$ is a small constant for numerical stability.
In implementation, we stop gradients on the semantic feature when computing $\mathcal{L}_{\mathrm{orth}}$, so that this constraint mainly updates the style branch and prevents the semantic branch from being overly disturbed. In this way, the style branch is encouraged to encode domain-related appearance factors that are complementary, rather than redundant, to the semantic counting representation.

\subsection{Granular Ball Guided Stable Latent Domain Discovery}
\label{sec:gb_pseudodomain}

\begin{figure*}[H]
    \centering
    \includegraphics[width=\linewidth]{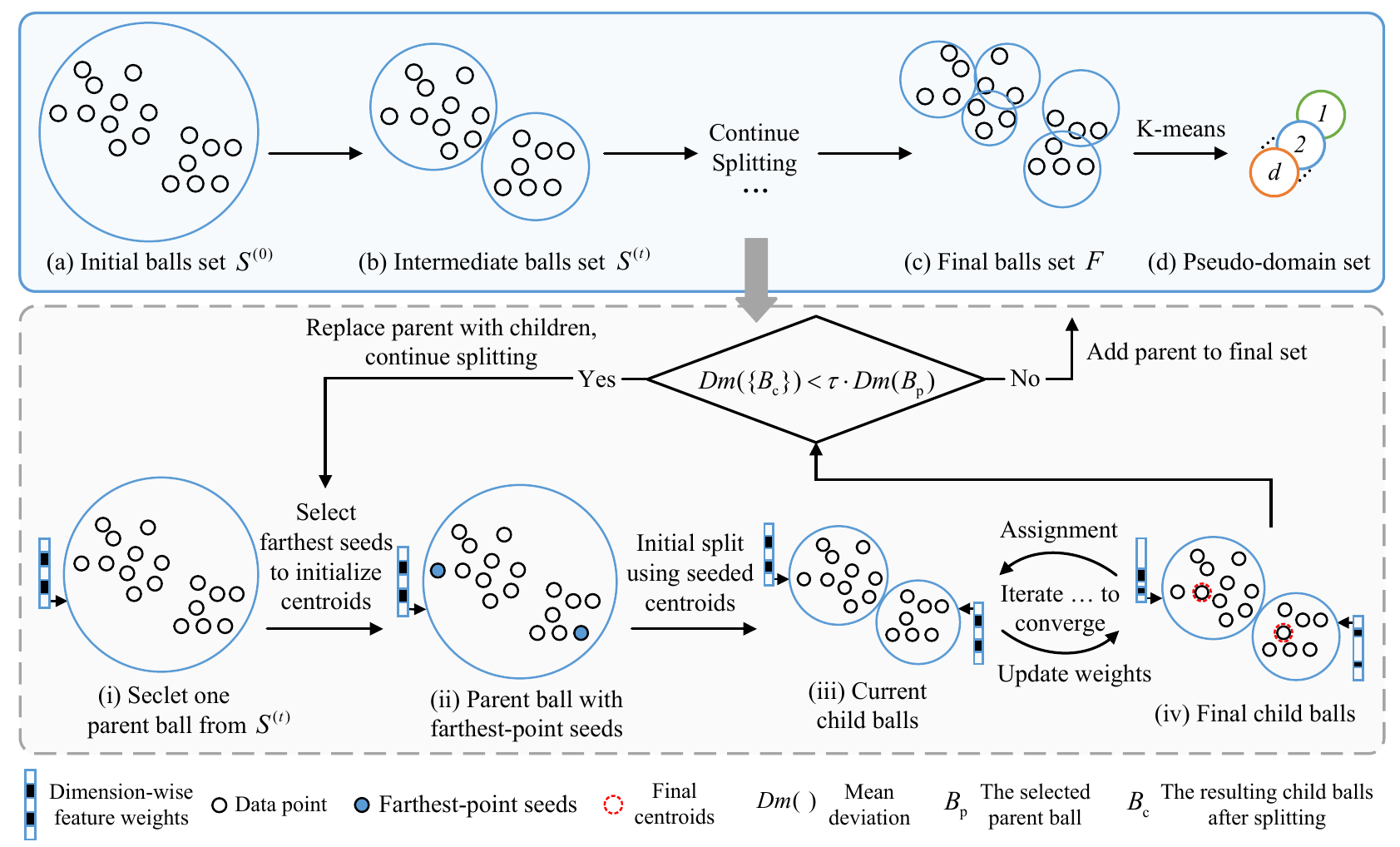}
    \caption{Hierarchical division and merging of a weighted granular ball set (WGBS) for pseudo-domain discovery. Starting from the initial ball set $S^{(0)}$, parent balls are iteratively split by weighted 2-means and accepted only when the child compactness decreases sufficiently. After no further split is accepted, $K$-means is performed on granular ball centers to obtain pseudo-domains.}
    \label{fig:fragmentation}
\end{figure*}

A key challenge in single-source domain generalization is how to stably discover latent domain structure from evolving sample representations. Existing pseudo-domain discovery methods typically perform flat clustering directly on sample-level features, which is vulnerable to noisy samples, outliers, and representation drift during training. To address this issue, we propose a granular ball guided stable latent domain discovery strategy. Instead of clustering all samples directly, we first organize the domain-sensitive descriptors into compact local granular balls, and then cluster granular ball centers as representatives to obtain $K$ pseudo-domains. In this way, direct sample-level clustering is transformed into a hierarchical representative-based clustering process, yielding more stable pseudo-domain assignments under a single labeled source domain. The overall procedure is summarized in Algorithm~\ref{alg:gb_pseudodomain_short}.

\smallskip
At each update stage (e.g., once per epoch), we extract a domain-sensitive descriptor for every source sample. Given a source image $x_i$, the backbone produces intermediate feature maps $\{F_i^{(l)}\}_{l=1}^{L}$. We compute channel-wise statistics at each level as
\begin{equation}
\mu_i^{(l)} = \mathrm{Mean}(F_i^{(l)}), \qquad
\sigma_i^{(l)} = \mathrm{Std}(F_i^{(l)}),
\end{equation}
and concatenate them into an instance descriptor
\begin{equation}
z_i = [\mu_i^{(1)},\sigma_i^{(1)},\ldots,\mu_i^{(L)},\sigma_i^{(L)}] \in \mathbb{R}^{D}.
\end{equation}
These statistics are sensitive to appearance and style variations, and are therefore suitable for discovering latent domains. We further apply PCA to obtain a reduced descriptor $\tilde{z}_i \in \mathbb{R}^{d}$, and denote the reduced descriptor set by
\begin{equation}
\tilde{\mathcal{Z}}=\{\tilde{z}_i\}_{i=1}^{N}.
\end{equation}

\smallskip
A granular ball $B \subset \tilde{\mathcal{Z}}$ is defined as a compact subset of descriptors with center
\begin{equation}
c_B = \frac{1}{|B|}\sum_{\tilde{z}\in B}\tilde{z}.
\end{equation}
To improve splitting flexibility in the reduced feature space, we associate each ball with a non-negative feature-weight vector $w_B \in \mathbb{R}^{d}$ satisfying $\sum_{j=1}^{d} w_{B,j}=1$, and define the weighted distance between a descriptor $\tilde{z}$ and a center $c$ as
\begin{equation}
d_{w_B}(\tilde{z},c)
=
\sqrt{\sum_{j=1}^{d} w_{B,j}\,(\tilde{z}_j-c_j)^2 }.
\end{equation}

\smallskip
Starting from the root ball containing all descriptors, we recursively split each candidate ball into two child balls via weighted $2$-means. For each split attempt, we initialize the two centroids by farthest-point seeding within the parent ball, perform an initial assignment, and then iterate assignment, centroid update, and weight update until convergence. Given a ball $B$, we solve
\begin{equation}
\begin{aligned}
\min_{\{u_{i\ell}\},\{\theta_\ell\},w}\quad
& \sum_{\ell=1}^{2}\sum_{\tilde{z}_i\in B}\sum_{j=1}^{d}
u_{i\ell}\, w_j^{\beta}\,(\tilde{z}_{i,j}-\theta_{\ell,j})^2, \\
\text{s.t.}\quad
& \sum_{\ell=1}^{2}u_{i\ell}=1,\qquad u_{i\ell}\in\{0,1\}, \\
& w_j \ge 0,\qquad \sum_{j=1}^{d} w_j = 1,
\end{aligned}
\label{eq:wkmeans_obj}
\end{equation}
where $\theta_\ell\in\mathbb{R}^{d}$ denotes the $\ell$-th centroid and $\beta>1$ controls the sharpness of the feature weights. We optimize Eq.~\eqref{eq:wkmeans_obj} by alternating updates of assignments $\{u_{i\ell}\}$, centroids $\{\theta_\ell\}$, and weights $w$. Let
\begin{equation}
D_j = \sum_{\ell=1}^{2}\sum_{\tilde{z}_i\in B} u_{i\ell}\,(\tilde{z}_{i,j}-\theta_{\ell,j})^2
\end{equation}
denote the within-cluster scatter on dimension $j$. The feature weights are updated by inverse-scatter normalization:
\begin{equation}
w_j =
\begin{cases}
0, & D_j = 0,\\[2pt]
\displaystyle
\left(\sum_{t=1}^{d}
\left(\frac{D_j+\epsilon}{D_t+\epsilon}\right)^{\frac{1}{\beta-1}}
\right)^{-1},
& D_j \neq 0,
\end{cases}
\label{eq:w_update}
\end{equation}
where $\epsilon$ is a small constant. Intuitively, dimensions with smaller within-ball scatter receive larger weights, which leads to more discriminative local partitioning.

\smallskip
Not every split is accepted. We measure the compactness of a ball $B$ by the mean weighted deviation
\begin{equation}
\mathrm{Dm}(B;w_B)
=
\frac{1}{|B|}\sum_{\tilde{z}\in B} d_{w_B}(\tilde{z},c_B).
\label{eq:dm_ball}
\end{equation}
Suppose splitting $B$ produces two child balls $B_1$ and $B_2$ with a shared weight vector $w$ learned from Eq.~\eqref{eq:wkmeans_obj}. Their size-weighted compactness is computed as
\begin{equation}
\mathrm{Dm}_{\mathrm{child}}
=
\frac{|B_1|\mathrm{Dm}(B_1;w)+|B_2|\mathrm{Dm}(B_2;w)}{|B_1|+|B_2|}.
\end{equation}
We accept the split if
\begin{equation}
\mathrm{Dm}_{\mathrm{child}} < \tau\,\mathrm{Dm}(B;w_B),
\label{eq:split_rule}
\end{equation}
where $\tau$ is a fixed split margin. If the condition holds, the parent ball is replaced by its two children and the recursive splitting continues; otherwise, the parent ball is kept as a final ball. We repeat this divide-and-test procedure until no further split is accepted, or a predefined maximum depth is reached, resulting in a final granular ball set
\begin{equation}
\mathcal{B}=\{B_m\}_{m=1}^{M_b}.
\end{equation}

\smallskip
After hierarchical division, we compute a representative center for each granular ball:
\begin{equation}
g_m=\frac{1}{|B_m|}\sum_{\tilde{z}\in B_m}\tilde{z}.
\end{equation}
Instead of directly clustering all sample descriptors, we run standard $K$-means on the granular ball centers $\{g_m\}_{m=1}^{M_b}$ to obtain $K$ pseudo-domains. Each sample then inherits the pseudo-domain label of the granular ball it belongs to. In this way, the final pseudo-domain partition is derived from representative-level clustering rather than sample-level flat clustering. When the number of valid granular balls is insufficient (e.g., $M_b<K$), we directly apply $K$-means to all descriptors $\tilde{\mathcal{Z}}$ as a fallback.

\smallskip
Since pseudo-domains are re-discovered online, cluster indices may permute across epochs. To maintain label consistency over training, we align the newly obtained pseudo-domain labels with those from the previous epoch using maximum-overlap assignment (Hungarian matching), and reindex the current clusters accordingly. The aligned pseudo-domain labels are then written back to the dataset and used for domain-structured training in the subsequent epoch.

\begin{algorithm}[!t]
\caption{Granular ball guided stable latent domain discovery}
\label{alg:gb_pseudodomain_short}
\begin{algorithmic}[1]
\Require Descriptors $\{z_i\}_{i=1}^{N}$; PCA dimension $d$; pseudo-domain number $K$; split margin $\tau$; weight sharpness $\beta$; maximum depth $D_{\max}$; $\epsilon$; (optional) previous labels $\hat d^{\,\mathrm{prev}}$.
\Ensure Pseudo-domain labels $\{\hat d_i\}_{i=1}^{N}$.
\State $\tilde z_i \gets \mathrm{PCA}(z_i)\in\mathbb{R}^{d}$ for all $i$; initialize the root ball $B_0=\{\tilde z_i\}$ with $w_{B_0}=\frac{1}{d}\mathbf{1}$.
\State Initialize queue $\mathcal{Q}\!\leftarrow\!\{(B_0,0)\}$ and leaf set $\mathcal{B}\!\leftarrow\!\emptyset$.
\While{$\mathcal{Q}$ is not empty}
  \State Pop $(B,dep)$ from $\mathcal{Q}$.
  \State $(B_1,B_2,w)\gets$ weighted $2$-means on $B$ (farthest-point initialization; alternating updates of assignments, centroids, and weights using Eq.~\eqref{eq:wkmeans_obj} and Eq.~\eqref{eq:w_update}).
  \State Compute $\mathrm{Dm}(B_1;w)$, $\mathrm{Dm}(B_2;w)$, and $\mathrm{Dm}_{\mathrm{child}}$ using Eq.~\eqref{eq:dm_ball}.
  \If{$\mathrm{Dm}_{\mathrm{child}} < \tau\,\mathrm{Dm}(B;w_B)$ \textbf{and} $dep<D_{\max}$}
     \State Push $(B_1,dep{+}1)$ and $(B_2,dep{+}1)$ into $\mathcal{Q}$ with $w_{B_1}=w_{B_2}=w$.
  \Else
     \State $\mathcal{B}\leftarrow \mathcal{B}\cup\{B\}$.
  \EndIf
\EndWhile
\If{$|\mathcal{B}|<K$}
  \State Run $K$-means on $\{\tilde z_i\}_{i=1}^{N}$ to obtain $\{\hat d_i\}$.
\Else
  \State Compute granular ball centers $g_m$ for all $B_m\in\mathcal{B}$.
  \State Run $K$-means on $\{g_m\}$ to obtain $K$ pseudo-domains.
  \State Assign each sample the label of the granular ball it belongs to, yielding $\{\hat d_i\}$.
\EndIf
\If{$\hat d^{\,\mathrm{prev}}$ is available}
  \State Align current labels to $\hat d^{\,\mathrm{prev}}$ by Hungarian matching and reindex $\{\hat d_i\}$.
\EndIf
\State \Return $\{\hat d_i\}_{i=1}^{N}$.
\end{algorithmic}
\end{algorithm}

\subsection{Time Complexity of Granular Ball Division}
\label{sec:gb_time}

We analyze the computational complexity of the hierarchical granular ball division used in stable latent domain discovery. Let $N$ denote the number of descriptors, $d$ the reduced feature dimension after PCA, and $h_{\max}$ the maximum division depth. Consider a granular ball $B$ containing $n$ descriptors. To split $B$, we perform weighted 2-means with alternating updates of sample assignments, cluster centers, scatters $\{D_j\}$, and feature weights $w$. Each such iteration requires $\mathcal{O}(nd)$ operations, while the compactness evaluation in Eq.~\eqref{eq:dm_ball} has the same order. Therefore, one split attempt on $B$ costs $\mathcal{O}(Tnd)$, where $T$ is the number of alternating iterations. At any fixed depth of the hierarchy, the candidate granular balls form a partition of the descriptor set, and thus the sum of their sample sizes is $N$. Therefore, the total cost at one depth is $\mathcal{O}(TNd)$. Since the recursive division proceeds for at most $h_{\max}$ levels, the overall complexity of the granular ball division stage is bounded by $\mathcal{O}(TNd\,h_{\max})$. Under fixed implementation settings, where $d$, $T$, and $h_{\max}$ are treated as constants, the cost of the granular ball division stage grows linearly with the number of descriptors $N$.

In practice, the split criterion in Eq.~\eqref{eq:split_rule} often stops the division early for many granular balls, so the average runtime is typically lower than this worst-case upper bound. After hierarchical division, the subsequent $K$-means clustering is performed on $M_b$ granular ball centers rather than on all $N$ descriptors, with complexity $\mathcal{O}(IKM_b d)$, where $I$ is the number of $K$-means iterations and $K$ is the number of pseudo-domains. Since $M_b \ll N$ in practice, this representative-level clustering introduces only a relatively small additional overhead compared with direct sample-level clustering.

\subsection{Training Objective and Optimization}
\label{sec:opt}

We train the network by combining the density regression loss with the proposed domain-structured regularizers:
\begin{equation}
\mathcal{L}
= \mathcal{L}_{\mathrm{den}}
+ \lambda_{\mathrm{sem}} \mathcal{L}_{\mathrm{sem}}
+ \lambda_{\mathrm{sty}} \mathcal{L}_{\mathrm{sty}}
+ \lambda_{\mathrm{orth}} \mathcal{L}_{\mathrm{orth}}.
\label{eq:overall_loss}
\end{equation}
We supervise density estimation by a pixel-wise regression loss:
\begin{equation}
\mathcal{L}_{\mathrm{den}}
= \frac{1}{B}\sum_{i=1}^{B}\left\|\hat{y}_i - y_i\right\|_2^2,
\end{equation}
where $\hat{y}_i$ and $y_i$ denote the predicted and ground-truth density maps of the $i$-th image, respectively. The remaining terms encourage semantic consistency across the discovered latent domains, intra-domain style compactness, and semantic--style disentanglement, respectively. 

Notably, granular ball guided stable latent domain discovery is not introduced as an additional differentiable loss term. Instead, it is performed online to update pseudo-domain labels, which in turn provide the structural supervision required by $\mathcal{L}_{\mathrm{sem}}$ and $\mathcal{L}_{\mathrm{sty}}$. In this way, stable latent domain discovery and representation learning are coupled through iterative pseudo-domain refinement during training.

\section{Experiments}
\label{sec:exp}

\subsection{Datasets}
\label{sec:datasets}

We conduct experiments on four widely-used crowd counting benchmarks: ShanghaiTech Part A~\cite{zhang2016single} (SHA), ShanghaiTech~\cite{zhang2016single} Part B (SHB), UCF\_QNRF~\cite{idrees2018composition} (QNRF), and NWPU-Crowd~\cite{wang2020nwpu} (NWPU).
SHA contains 482 images with diverse resolutions and heavy congestion (501 persons per image on average), while SHB is comparatively sparse with a resolution of $768\times1024$ and 123 persons per image on average.
QNRF contains 1,535 images with 1,251,642 head annotations, where 1,201 images are used for training and 334 for testing.
NWPU is a large-scale dataset comprising 5,109 images annotated with 2.13 million points; it provides 3,109/500/1,500 images for train/val/test splits, respectively, and the test annotations are hidden (evaluation is performed via the official server).

\subsection{Experimental Setting}
\label{sec:exp_setting}

Following the single-source domain generalization protocol, the model is trained on one labeled source dataset and directly evaluated on the other unseen datasets without any target-domain fine-tuning or adaptation. 
Head-center annotations are transformed into density maps by Gaussian convolution with a fixed kernel size of 15. 
For fair comparison, we adopt the same backbone as DGCC~\cite{du2023domain}, i.e., an encoder--decoder architecture equipped with a $1\times1$ convolutional density estimator. 
Unless otherwise specified, training is performed for 200 epochs.
During training, input images are randomly cropped and resized to $320\times320$, and random horizontal flipping is applied for data augmentation. 
The Adam optimizer is used with an initial learning rate of $1\times10^{-5}$ and a weight decay of $1\times10^{-4}$. 
All experiments are conducted on a single NVIDIA A100-SXM-64GB GPU with a batch size of 32. 
For stable latent domain discovery, the number of pseudo-domains $K$ is determined according to the source dataset, and the pseudo-domain labels are updated online at the end of each epoch. 
Specifically, $K$ is set to 4, 3, and 6 when the source dataset is SHA, SHB, and QNRF, respectively. 
For granular ball division, the split margin is fixed at $\tau{=}1.05$ across all experiments. 
Following prior work, performance is evaluated using mean absolute error (MAE) and mean squared error (MSE).

\begin{table*}[!t]
\caption{Comparison with state-of-the-art no-adaptation methods under different source$\rightarrow$target cross-domain transfer settings (MAE/MSE). Venue denotes the publication venue of each method. Results are grouped by the source training set (SHA, SHB, or QNRF) and evaluated on the remaining unseen datasets. Bold values denote the best results.}
\label{tab:sota_all}
\centering
\small
\setlength{\tabcolsep}{4.2pt}
{\renewcommand{\arraystretch}{1.18}
\resizebox{\textwidth}{!}{
\begin{tabular}{l c cc cc cc cc cc cc}
\toprule
Source &  &
\multicolumn{4}{c}{SHA} &
\multicolumn{4}{c}{SHB} &
\multicolumn{4}{c}{QNRF} \\
\cmidrule(lr){3-6}\cmidrule(lr){7-10}\cmidrule(lr){11-14}
Target & &
\multicolumn{2}{c}{SHB} & \multicolumn{2}{c}{QNRF} &
\multicolumn{2}{c}{SHA} & \multicolumn{2}{c}{QNRF} &
\multicolumn{2}{c}{SHA} & \multicolumn{2}{c}{SHB} \\
\cmidrule(lr){3-4}\cmidrule(lr){5-6}\cmidrule(lr){7-8}\cmidrule(lr){9-10}\cmidrule(lr){11-12}\cmidrule(lr){13-14}
Method & Publication & MAE & MSE & MAE & MSE & MAE & MSE & MAE & MSE & MAE & MSE & MAE & MSE \\
\midrule
DMCount~\cite{wang2020distribution}        & NeurIPS'20          & 23.1 & 34.9  & 134.4 & 252.1 & 143.9 & 239.6 & 203.0 & 386.1 & 73.4 & 135.1 & 14.3 & 27.5 \\
D2CNet~\cite{cheng2021decoupled}           & TIP'21              & 21.6 & 34.6  & 126.8 & 245.5 & 164.5 & 286.4 & 267.5 & 486.0 & --   & --    & --   & -- \\
SASNet~\cite{song2021choose}               & AAAI'21             & 21.3 & 33.2  & 211.2 & 418.6 & 132.4 & 225.6 & 273.5 & 481.3 & 73.9 & 116.4 & 13.0 & 22.1 \\
MAN~\cite{lin2022boosting}                 & CVPR'22             & 22.1 & 32.8  & 138.8 & 266.3 & 133.6 & 255.6 & 209.4 & 378.8 & 67.1 & 122.1 & 12.5 & 22.2 \\
DG-MAN~\cite{lin2022boosting,du2023domain} & AAAI'23             & 17.3 & 28.7  & 129.1 & 238.2 & 130.7 & 225.1 & 182.4 & 325.8 & --   & --    & --   & -- \\
DGCC~\cite{du2023domain}                   & AAAI'23             & 12.6 & 24.6  & 119.4 & 216.6 & 121.8 & 203.1 & 179.1 & 316.2 & 67.4 & 112.8 & 12.1 & 20.9 \\
MPCount~\cite{peng2024single}              & CVPR'24             & 11.4 & 19.7  & 115.7 & 199.8 & 99.6  & 182.9 & 165.6 & 290.4 & 65.5 & 110.1 & 12.3 & 24.1 \\
UGSDA~\cite{ding2024domain}                & ACM MM'24           & 11.6 & 24.5  & 117.0 & 194.1 & 113.4 & 180.8 & 178.1 & 306.7 & 65.8 & 104.0 & \textbf{10.9} & \textbf{19.1} \\
MAHE~\cite{zhou2025crowd}                  & Sci. Rep.'25         & 12.1 & 19.5  & 118.1 & 203.5 & 103.6 & 178.7 & 162.5 & \textbf{285.8} & 67.3 & 108.7 & 12.2 & 20.7 \\
SinCount~\cite{song2025sincount}           & Res. Sq.'25  & 11.8 & 20.0  & 109.0 & \textbf{180.0} & 104.9 & 186.1 & 169.2 & 298.1 & 73.8 & 125.3 & 11.9 & 24.1 \\
ADG~\cite{zhang2026adversarial}            & MTAP'26             & 12.8 & 25.8  & 115.3 & 200.2 & 98.5  & 175.5 & 164.5 & 286.7 & \textbf{64.1} & 108.2 & 13.1 & 25.0 \\
\midrule
GBDGC (Ours)                               & --                  & \textbf{11.1} & \textbf{15.7} & \textbf{108.0} & 197.4 & \textbf{97.5} & \textbf{171.7} & \textbf{160.3} & 298.8 & \textbf{64.1} & \textbf{103.5} & 13.0 & 21.5 \\
\bottomrule
\end{tabular}}
}
\end{table*}

\subsection{Comparison with State of the Art}
\label{sec:exp_sota}

Table~\ref{tab:sota_all} reports the cross-domain comparison results under the single-source, no-adaptation protocol. Overall, GBDGC demonstrates strong generalization across diverse cross-domain transfer settings. For the transfer settings SHA$\rightarrow$SHB, SHA$\rightarrow$QNRF, SHB$\rightarrow$SHA, and SHB$\rightarrow$QNRF, GBDGC achieves the lowest MAE in all four cases, indicating consistently improved counting accuracy under different source--target combinations. In terms of MSE, it obtains the best results on SHA$\rightarrow$SHB and SHB$\rightarrow$SHA, while remaining competitive on SHA$\rightarrow$QNRF and SHB$\rightarrow$QNRF. These results suggest that the proposed method is effective not only under relatively moderate domain shifts, but also under more challenging transfer scenarios involving substantial differences in density range, scene layout, and imaging conditions. Its advantage is particularly evident on SHB$\rightarrow$QNRF, where it achieves the best MAE of 160.3 with competitive MSE. This result is notable because QNRF contains highly diverse crowd scenes and large count variations, making cross-dataset generalization especially difficult. We further examine reverse transfer settings by using QNRF as the source domain and ShanghaiTech as the target domain. Under QNRF$\rightarrow$SHA, GBDGC achieves a tie for the best MAE (64.1) and the best MSE (103.5), indicating strong transferability from a highly diverse source domain. This observation further suggests that the proposed latent domain modeling strategy can exploit the structural diversity of the source dataset and translate it into more robust representations for unseen domains. On QNRF$\rightarrow$SHB, although it does not achieve the top result, GBDGC still yields competitive performance of 13.0/21.5 (MAE/MSE), further demonstrating its robustness under severe cross-dataset distribution shifts. Taken together, these results show that GBDGC maintains stable performance across both conventional and reverse transfer settings, supporting its effectiveness as a general framework for single-source domain-general crowd counting.

We further evaluate GBDGC on the large-scale NWPU-Crowd benchmark to assess its out-of-domain generalization under more challenging scene diversity and count variation. As reported in Table~\ref{tab:sota_nwpu}, GBDGC consistently achieves the lowest MSE across all three training settings, indicating improved robustness with fewer extreme prediction errors. This trend is particularly meaningful on NWPU, where scenes are highly heterogeneous and the count distribution is long-tailed, making models more susceptible to substantial overestimation or underestimation. In particular, when trained on SHB, GBDGC improves both metrics over DGCC~\cite{du2023domain}, reducing MAE from 175.0 to 173.4 and MSE from 688.6 to 598.5, which suggests a clear reduction in large-count outliers. When trained on SHA or SHA+SHB, GBDGC also produces consistent MSE reductions, for example from 567.6 to 546.2 and from 553.6 to 545.0, while maintaining comparable MAE. Overall, these results indicate that GBDGC not only improves average counting accuracy but also yields more reliable predictions under challenging out-of-domain conditions. For the joint-source setting SHA+SHB, the number of pseudo-domains is set to $K{=}5$.

\begin{table}[H]
\caption{Additional evaluation on NWPU-Crowd under a no-adaptation setting. Models are trained on SHA, SHB, or their union, and directly tested on NWPU.}
\label{tab:sota_nwpu}
\centering
\small
\setlength{\tabcolsep}{5.0pt}
{\renewcommand{\arraystretch}{1.20}
\resizebox{0.48\textwidth}{!}{
\begin{tabular}{l cc cc cc}
\toprule
Source
& \multicolumn{2}{c}{SHA}
& \multicolumn{2}{c}{SHB}
& \multicolumn{2}{c}{SHA+SHB} \\
\cmidrule(l{1pt}r{30pt}){1-1}\cmidrule(lr){2-3}\cmidrule(lr){4-5}\cmidrule(lr){6-7}
Target
& \multicolumn{6}{c}{NWPU} \\
\cmidrule(l{1pt}r{30pt}){1-1}\cmidrule(lr){2-7}
Method
& MAE & MSE & MAE & MSE & MAE & MSE \\
\midrule
DMCount~\cite{wang2020distribution}  & 146.9 & 563.8 & 191.6 & 747.4 & 144.6 & 592.8 \\
SASNet~\cite{song2021choose}         & 158.8 & 588.0 & 195.7 & 716.8 & 155.3 & 583.6 \\
MAN~\cite{lin2022boosting}           & 148.2 & 586.5 & 193.6 & 802.5 & 147.8 & 605.3 \\
DGCC~\cite{du2023domain}             & \textbf{143.1} & 567.6 & 175.0 & 688.6 & \textbf{139.6} & 553.6 \\
\midrule
GBDGC (Ours) & 144.3 & \textbf{546.2} & \textbf{173.4} & \textbf{598.5} & 144.1 & \textbf{545.0} \\
\bottomrule
\end{tabular}}
}
\end{table}

\begin{figure*}[!htp]
    \centering
    \includegraphics[width=\textwidth]{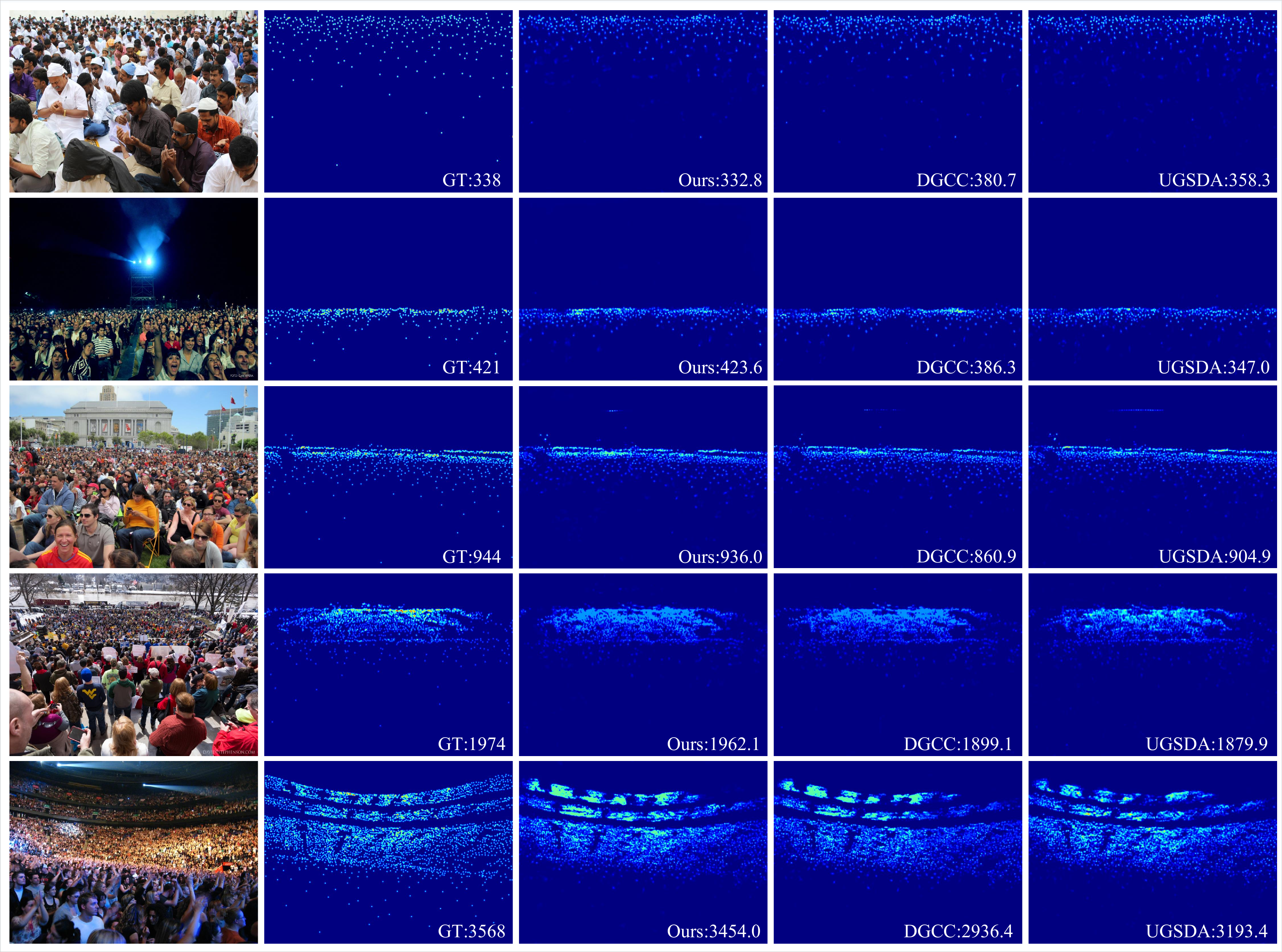}
    \caption{Qualitative comparison on UCF\_QNRF under the no-adaptation setting.
    From left to right: input image, GT density map, and the predictions of Ours, DGCC, UGSDA.
    The numbers denote crowd counts obtained by integrating the density maps.}
    \label{fig:qual_qnrf}
\end{figure*}

Fig.~\ref{fig:qual_qnrf} presents qualitative comparisons on the unseen QNRF target domain without target-domain adaptation. For a fair comparison, all methods are initialized from the publicly released SHA pre-trained weights provided in their official implementations, and our visualizations are also generated using the SHA pre-trained model only. The target scenes exhibit clear domain shifts in appearance and imaging conditions, including large perspective variations, complex illumination, and wide density ranges. Under these challenging conditions, our method produces density maps that are visually closer to the ground truth, with responses more accurately concentrated on true crowd regions and with better-preserved global spatial layouts. In particular, our predictions remain more coherent across both sparse and highly congested areas, while being less affected by over-smoothing and background interference. By contrast, DGCC~\cite{du2023domain} and UGSDA~\cite{ding2024domain} tend to show weakened responses in distant dense regions, blurrier density distributions, and occasional spurious activations caused by background textures or lighting variations.

\subsection{Ablation Study}
\subsubsection{Effect of Pseudo-domain Partitioning}

We study the impact of pseudo-domain partitioning, focusing on the partitioning strategy rather than the choice of the number of pseudo-domains.
For a clean ablation, we fix the number of pseudo-domains and discuss the selection of $K$ in the next subsection.
Following the settings used in prior work, we set $K{=}4$ on SHA, $K{=}3$ on SHB, and $K{=}6$ on QNRF, while keeping the backbone, training schedule, and all loss weights unchanged.
We only vary how pseudo-domain labels are assigned:
(i) Random partition, which uniformly shuffles and splits the training set into $K$ groups as a structure-free sanity check;
(ii) K-means partition, which follows the clustering baseline used in~\cite{du2023domain};
and (iii) Ours, which performs granular ball guided partitioning and then merges clusters into $K$ groups.
Table~\ref{tab:ablation_domain_partition} summarizes the results. Under the same fixed $K$, our partitioning achieves the best MAE across all six source$\rightarrow$target settings, indicating that pseudo-domain construction remains crucial even when the number of groups is held constant.
In terms of MSE, our method obtains the best results in five out of six settings and remains competitive on the remaining transfer.
Compared with Random grouping, our strategy brings consistent improvements, with a particularly clear gain on the challenging QNRF$\rightarrow$SHA transfer (64.1/103.5 vs.\ 68.6/114.6).
Compared with the K-means baseline~\cite{du2023domain}, our method consistently improves MAE in all settings and also improves MSE in most settings, with larger gains under more severe domain shifts.
Although Random grouping can be competitive in a few cases, its behavior is less stable overall, suggesting that the improvements do not come from arbitrary grouping, but from a more informative and structure-aware partitioning strategy.

Beyond transfer accuracy, we further investigate whether the discovered pseudo-domains align with a meaningful semantic factor in crowd counting, namely the count (density) regime. This factor also helps explain why a better partition improves transferability: different density regimes exhibit distinct occlusion patterns, textures, and feature statistics, and pseudo-domain learning is more effective when each group is internally coherent rather than mixing sparse and highly crowded samples. A random split tends to average out regime-specific statistics and thus weakens the pseudo-domain signal, whereas a regime-aware partition produces more homogeneous groups and a more informative regularization signal across domains, consistent with the transfer gains in Table~\ref{tab:ablation_domain_partition}. Using the same fixed $K$ as in Table~\ref{tab:ablation_domain_partition}, we summarize each pseudo-domain $d$ by its median ground-truth count $m_d$, and evaluate how well a partition separates density regimes by measuring the spread of these $K$ medians, namely $\Delta_{\mathrm{med}}=\max(m_d)-\min(m_d)$ and the standard deviation $\sigma_{\mathrm{med}}$ over $\{m_d\}_{d=1}^{K}$ (Table~\ref{tab:count_separation}; larger values indicate better separation). If a partition is close to random, each group tends to reflect the global count distribution, resulting in similar medians and therefore small $\Delta_{\mathrm{med}}$ and $\sigma_{\mathrm{med}}$; by contrast, larger values indicate that the pseudo-domains are organized more clearly along the density axis, with less mixing between sparse and crowded samples. Table~\ref{tab:count_separation} shows that Random grouping yields consistently weak separation, while K-means already enlarges the spread of pseudo-domain medians. Our method further increases both $\Delta_{\mathrm{med}}$ and $\sigma_{\mathrm{med}}$ on all datasets. The effect is most pronounced on the long-tailed QNRF dataset, where $\Delta_{\mathrm{med}}$ increases from 1624.7 to 1949.6 and $\sigma_{\mathrm{med}}$ increases from 563.3 to 686.6, indicating a clearer isolation of extreme-density regimes.

\begin{table*}[]
\caption{Ablation on pseudo-domain partitioning (MAE/MSE; lower is better).
We fix $K$ for a clean module ablation ($K{=}4$ on SHA, $K{=}3$ on SHB, and $K{=}6$ on UCF\_QNRF); the choice of $K$ is discussed in the next subsection.
Best results are highlighted in boldface.}
\label{tab:ablation_domain_partition}
\centering
\fontsize{7}{8}\selectfont
\setlength{\tabcolsep}{2.5pt}
\renewcommand{\arraystretch}{1.05}
\resizebox{0.9\textwidth}{!}{
\begin{tabular}{@{}l cc cc cc cc cc cc@{}}
\toprule
Source
& \multicolumn{4}{c}{SHA}
& \multicolumn{4}{c}{SHB}
& \multicolumn{4}{c}{UCF\_QNRF} \\
\cmidrule(lr){2-5}\cmidrule(lr){6-9}\cmidrule(lr){10-13}
Target
& \multicolumn{2}{c}{SHB} & \multicolumn{2}{c}{QNRF}
& \multicolumn{2}{c}{SHA} & \multicolumn{2}{c}{QNRF}
& \multicolumn{2}{c}{SHA} & \multicolumn{2}{c}{SHB} \\
\cmidrule(lr){2-3}\cmidrule(lr){4-5}\cmidrule(lr){6-7}\cmidrule(lr){8-9}\cmidrule(lr){10-11}\cmidrule(lr){12-13}
Method
& MAE & MSE & MAE & MSE & MAE & MSE & MAE & MSE & MAE & MSE & MAE & MSE \\
\midrule
Random partition
& 12.8 & 23.4 & 119.4 & 218.4
& 120.2 & 210.2 & 169.2 & 307.1
& 68.6 & 114.6 & 14.3 & 27.0 \\
K-means
& 12.9 & 24.4 & 119.8 & 219.6
& 122.7 & 220.2 & 166.1 & \textbf{295.3}
& 68.0 & 111.3 & 13.5 & 25.3 \\
\midrule
Ours
& \textbf{11.1} & \textbf{15.7} & \textbf{108.0} & \textbf{197.4}
& \textbf{97.5} & \textbf{171.7} & \textbf{160.3} & 298.8
& \textbf{64.1} & \textbf{103.5} & \textbf{13.0} & \textbf{21.5} \\
\bottomrule
\end{tabular}
}
\end{table*}

\begin{table}[!t]
\caption{GT-count stratification of pseudo-domains.
For each method, we compute the median GT count $m_d$ for each pseudo-domain and report the range $\Delta_{\mathrm{med}}=\max(m_d)-\min(m_d)$ and the standard deviation $\sigma_{\mathrm{med}}$ across the $K$ medians.}
\label{tab:count_separation}
\centering
\small
\setlength{\tabcolsep}{6pt}
\renewcommand{\arraystretch}{1.1}
\resizebox{0.38\textwidth}{!}{
\begin{tabular}{l l cc}
\toprule
Dataset & Method & $\Delta_{\mathrm{med}}\uparrow$ & $\sigma_{\mathrm{med}}\uparrow$ \\
\midrule
\multirow{3}{*}{SHA}
& Random  & 90.8  & 33.8  \\
& K-means & 270.9 & 110.2 \\
& Ours    & \textbf{278.0} & \textbf{112.7} \\
\midrule
\multirow{3}{*}{SHB}
& Random  & 20.7  & 8.5   \\
& K-means & 92.9  & 41.8  \\
& Ours    & \textbf{110.2} & \textbf{46.3} \\
\midrule
\multirow{3}{*}{QNRF}
& Random  & 72.4   & 23.5  \\
& K-means & 1624.7 & 563.3 \\
& Ours    & \textbf{1949.6} & \textbf{686.6} \\
\bottomrule
\end{tabular}
}
\end{table}

\subsubsection{Effect of the Number of Pseudo-domains $K$}
\label{sec:ablation_k}

We study the sensitivity of our method to the number of pseudo-domains $K$.
Table~\ref{tab:ablation_k} reports the results on three source datasets using several nearby choices of $K$.
Overall, the performance exhibits a typical under-/over-partitioning behavior.
When $K$ is too small, heterogeneous samples are forced into the same pseudo-domain, leading to high intra-domain variance and weakened domain-structured regularization.
When $K$ is too large, the training set is fragmented into overly small and noisy groups, making intra-/inter-domain statistics less reliable.
As a result, intermediate values of $K$ provide a better trade-off and generally yield lower MAE/MSE.

To avoid tuning $K$ over a wide range when transferring to a new source dataset, we adopt a simple data-dependent rule to determine a reasonable scale. Given the number of training images $N$, we compute
\begin{equation}
K_0 = N^{\frac{1}{4}},
\label{eq:k_heuristic}
\end{equation}
and then evaluate only a few candidates in a small neighborhood around $K_0$. This heuristic increases sublinearly with $N$, balancing partition granularity and the number of samples within each pseudo-domain. As shown in Table~\ref{tab:ablation_k}, the best-performing values are consistently found near the scale suggested by Eq.~(\ref{eq:k_heuristic}): we use $K{=}4$ for SHA, $K{=}3$ for SHB, and $K{=}6$ for QNRF. The smaller optimum on SHB suggests a more concentrated source distribution, for which fewer pseudo-domains yield more stable group statistics. Overall, Eq.~(\ref{eq:k_heuristic}) narrows the search range and reduces tuning cost when applying our framework to a new source dataset.

\begin{table*}[!t]
\centering
\caption{Effect of the number of pseudo-domains $K$ under the single-source domain-generalization setting.}
\label{tab:ablation_k}

\begin{minipage}[t]{0.58\textwidth}
\raggedleft
\small
\setlength{\tabcolsep}{4pt}
{\renewcommand{\arraystretch}{1.12}
\resizebox{1.0\textwidth}{!}{
\begin{tabular}{ccccccccc}
\toprule
Source
& \multicolumn{4}{c}{SHA} & \multicolumn{4}{c}{SHB} \\
\cmidrule(lr){1-1}\cmidrule(lr){2-5}\cmidrule(lr){6-9}
Target
& \multicolumn{2}{c}{SHB} & \multicolumn{2}{c}{QNRF}
& \multicolumn{2}{c}{SHA} & \multicolumn{2}{c}{QNRF} \\
\cmidrule(lr){1-1}\cmidrule(lr){2-3}\cmidrule(lr){4-5}\cmidrule(lr){6-7}\cmidrule(lr){8-9}
$K$
& MAE & MSE & MAE & MSE & MAE & MSE & MAE & MSE \\
\midrule
3 & 12.8 & 24.3 & 120.9 & 218.1 & \textbf{97.5} & \textbf{171.7} & \textbf{160.3} & \textbf{298.8} \\
4 & \textbf{11.1} & \textbf{15.7} & \textbf{108.0} & \textbf{197.4} & 119.3 & 202.9 & 172.7 & 306.4 \\
5 & 12.6 & 23.7 & 118.3 & 215.6 & 117.4 & 200.7 & 177.6 & 322.0 \\
\bottomrule
\end{tabular}}
}
\end{minipage}
\hspace*{2pt}
\begin{minipage}[t]{0.39\textwidth}
\raggedright
\small
\setlength{\tabcolsep}{6pt}
{\renewcommand{\arraystretch}{1.12}
\resizebox{0.94\textwidth}{!}{
\begin{tabular}{ccccc}
\toprule
Source
& \multicolumn{4}{c}{QNRF} \\
\cmidrule(lr){1-1}\cmidrule(lr){2-5}
Target
& \multicolumn{2}{c}{SHA} & \multicolumn{2}{c}{SHB} \\
\cmidrule(lr){1-1}\cmidrule(lr){2-3}\cmidrule(lr){4-5}
$K$
& MAE & MSE & MAE & MSE \\
\midrule
5 & 67.6 & 109.4 & 14.0 & 25.8 \\
6 & \textbf{64.1} & \textbf{103.5} & \textbf{13.0} & \textbf{21.5} \\
7 & 70.8 & 113.3 & 14.3 & 27.0 \\
\bottomrule
\end{tabular}}
}
\end{minipage}
\end{table*}

\begin{table*}[!t]
\centering
\caption{Ablation study of the core components (SCR and SBR) and the SBR regularizers under the single-source domain-generalization setting (without target-domain adaptation).}
\label{tab:ablation_scr_sbr}

\normalsize
\setlength{\tabcolsep}{5.0pt}
\renewcommand{\arraystretch}{1.20}
\resizebox{0.9\textwidth}{!}{
\begin{tabular}{@{}lcccccccccccc@{}}
\toprule
Source & \multicolumn{4}{c}{SHA} & \multicolumn{4}{c}{SHB} & \multicolumn{4}{c}{QNRF} \\
\cmidrule(r{20pt}){1-1}\cmidrule(lr){2-5}\cmidrule(lr){6-9}\cmidrule(lr){10-13}
Target & \multicolumn{2}{c}{SHB} & \multicolumn{2}{c}{QNRF}
                  & \multicolumn{2}{c}{SHA} & \multicolumn{2}{c}{QNRF}
                  & \multicolumn{2}{c}{SHA} & \multicolumn{2}{c}{SHB} \\
\cmidrule(r{20pt}){1-1}\cmidrule(lr){2-3}\cmidrule(lr){4-5}\cmidrule(lr){6-7}\cmidrule(lr){8-9}\cmidrule(lr){10-11}\cmidrule(lr){12-13}
Method & MAE & MSE & MAE & MSE & MAE & MSE & MAE & MSE & MAE & MSE & MAE & MSE \\
\midrule
baseline
& 14.8 & 27.7 & 134.5 & 230.5 & 132.6 & 229.7 & 190.6 & 305.9 & 88.2 & 145.0 & 18.8 & 35.6 \\
+SCR
& 13.8 & 27.2 & 123.8 & 228.6 & 121.7 & 203.2 & 171.4 & 298.9 & 69.9 & 116.7 & 15.4 & 28.7 \\
+SBR (Ours)
& \textbf{11.1} & \textbf{15.7} & \textbf{108.0} & \textbf{197.4}
& \textbf{97.5} & \textbf{171.7} & \textbf{160.3} & \textbf{298.8}
& \textbf{64.1} & \textbf{103.5} & \textbf{13.0} & \textbf{21.5} \\
\midrule
w/o $\mathcal{L}_{sem}$
& 13.9 & 27.0 & 117.8 & 214.1 & 119.3 & 214.8 & 172.7 & 306.4 & 67.2 & 111.0 & 14.1 & 26.5 \\
w/o $\mathcal{L}_{sty}$
& 13.0 & 24.1 & 123.0 & 232.9 & 122.1 & 223.6 & 169.2 & 307.1 & 66.5 & 104.1 & 14.6 & 28.8 \\
w/o $\mathcal{L}_{orth}$
& 12.7 & 25.7 & 125.1 & 234.5 & 118.5 & 209.8 & 178.7 & 313.6 & 71.6 & 118.6 & 13.8 & 25.6 \\
\bottomrule
\end{tabular}
}
\end{table*}

\subsubsection{Effect of Core Components}
\label{sec:ablation_core_components}

We evaluate the contribution of our two core components, namely Semantic Codebook Re-encoding (SCR) and Structured Branch Regularization (SBR).
All experiments are conducted under the standard single-source domain-generalization setting, where the model is trained on one source dataset and directly tested on unseen target datasets without any target-domain adaptation.
Table~\ref{tab:ablation_scr_sbr} reports the MAE/MSE results over six transfer settings.
Starting from the baseline, adding SCR leads to consistent improvements across all transfers, suggesting that codebook-based re-encoding provides more stable semantic representations under domain shift.
For example, SCR reduces the MAE on SHA$\rightarrow$QNRF from 134.5 to 123.8 and improves the performance on QNRF$\rightarrow$SHB from 18.8/35.6 to 15.4/28.7 (MAE/MSE).
Building upon SCR, further introducing SBR substantially strengthens cross-domain generalization and achieves the best overall performance across all transfer settings, including the challenging QNRF$\rightarrow$SHA case, where the MAE is further reduced from 69.9 to 64.1.

We further analyze the effect of the three regularizers in SBR by removing them one at a time.
As shown in Table~\ref{tab:ablation_scr_sbr}, removing any of $\mathcal{L}_{sem}$, $\mathcal{L}_{sty}$, or $\mathcal{L}_{orth}$ consistently degrades the performance, indicating that these regularizers provide complementary benefits.
Specifically, $\mathcal{L}_{sem}$ encourages semantic consistency across pseudo-domains, $\mathcal{L}_{sty}$ promotes style compactness within each pseudo-domain, and $\mathcal{L}_{orth}$ enhances semantic--style disentanglement by suppressing feature leakage between the two branches.
Overall, SCR improves the stability of semantic representation, while SBR further enhances generalization by jointly modeling semantic invariance, style compactness, and branch disentanglement.

\section{Conclusion}
\label{sec:conclusion}

In this paper, we proposed Granular Ball Domain-General Counting (GBDGC), a single-source domain generalization framework for crowd counting that explicitly exploits the latent domain structure hidden in a source dataset. To achieve robust pseudo-domain discovery under evolving feature representations, we introduced a granular ball guided online discovery strategy that combines coarse-to-fine set partitioning with granular ball center clustering, producing more stable pseudo-domain assignments than direct sample-level flat clustering. Based on the discovered latent domains, we further developed a two-branch learning framework, in which the semantic branch improves transferable counting representations via semantic codebook re-encoding (SCR), while the style branch enhances domain-aware regularization through intra-domain compactness and semantic--style orthogonality. Extensive experiments on SHA, SHB, UCF\_QNRF, and NWPU under standard domain generalization settings verify the effectiveness of the proposed method and show strong generalization ability, especially in transfer scenarios with large domain gaps.

In future work, we will extend granular ball structure modeling to more challenging settings, such as multi-source domain generalization and continuously evolving deployment environments. It is also of interest to combine the proposed framework with stronger backbones, such as transformer-based encoders, and to investigate more fine-grained structural cues for pseudo-domain discovery, such as region-aware and density-aware descriptors. We believe these directions can further improve robustness in highly congested scenes and complex imaging conditions.


\bibliographystyle{unsrtnat}
\bibliography{ref}

\end{document}